\documentclass{article}
\usepackage{xcolor}
\PassOptionsToPackage{sort&compress}{natbib}
\PassOptionsToPackage{pdfversion=1.7}{hyperref}
\usepackage[final]{corl_2026}
\usepackage[T1]{fontenc}
\usepackage[utf8]{inputenc}
\usepackage{amsmath,amssymb}
\usepackage{booktabs,tabularx,array}
\usepackage{graphicx}
\graphicspath{{figures/}}
\usepackage{afterpage,float}
\usepackage{xurl}
\usepackage{tikz}
\usetikzlibrary{arrows.meta,positioning}
\newcommand{\method}{DreamVLA}
\newcommand{\expert}{\pi_{\mathrm{E}}}

\newcommand{\dataset}{\mathcal{D}}
\newcolumntype{Y}{>{\raggedright\arraybackslash}X}
\hypersetup{
  hypertexnames=false,
  pdftitle={DreamVLA: From Generated Human Motions to Humanoid Vision-Language-Action Policies},
  pdfauthor={Kevin Park, Izat Rasyad, Dvij Kalaria, Shankar Sastry}
}

\title{DreamVLA: From Generated Human Motions to\protect\\
Humanoid Vision-Language-Action Policies}
\author{%
  Kevin Park$^{1}$\thanks{Equal contribution.}\quad
  Izat Rasyad$^{2}$\footnotemark[1]\quad
  Dvij Kalaria$^{2}$\quad
  Shankar Sastry$^{2}$\\
  \normalfont $^{1}$Princeton University\quad $^{2}$University of California, Berkeley\\
  \normalfont\texttt{kp034@princeton.edu}\\
  \normalfont\texttt{\{izat,dvijk,sastry\}@berkeley.edu}
}

\begin{document}
\maketitle

\begin{abstract}
Training vision-language-action (VLA) policies typically depends on substantial robot demonstration data, making human teleoperation a costly and cumbersome bottleneck, particularly for whole-body humanoid interaction. Reinforcement learning in simulation offers an alternative for scaling this data, but designing task-specific rewards remains challenging.
To address this, we propose DreamVLA, which uses generated motion trajectories to guide reinforcement learning (RL) in simulation, converting them into reliable task execution trajectories. These trajectories train VLA models in simulation without new task-specific teleoperation data. Our central contribution is a reference-conditioned DAgger distillation procedure: after a failed learner rollout, we restart the RL expert from a sampled learner state with the same reference suffix and aggregate successful corrective trajectories, addressing the distribution shift left by offline imitation.
We evaluate picking from a platform, opening a drawer, pressing a button, and walking and picking from a platform, with ablations of DAgger iterations, residual RL, and generated dataset scale.
\\
\textbf{Project website:} \href{https://ember-center-berkeley.github.io/DreamVLA/}{\nolinkurl{ember-center-berkeley.github.io/DreamVLA}}.
\end{abstract}


\afterpage{%
\clearpage
\begin{figure}[H]
\centering
\includegraphics[width=\linewidth]{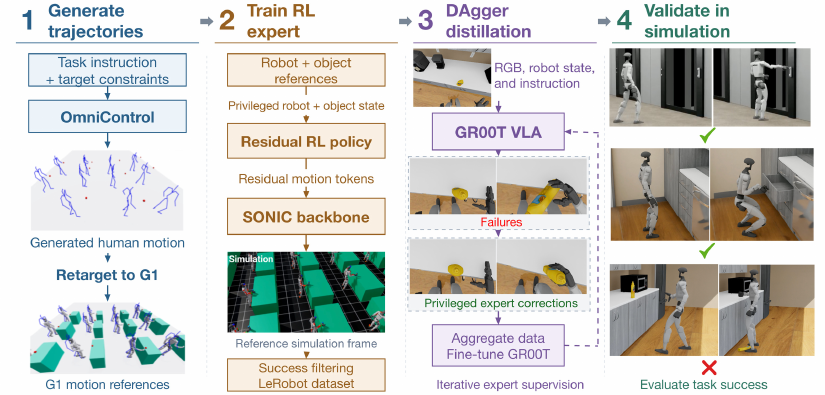}
\caption{\method{} overview. Task-conditioned human motions are generated with DreamControl/OmniControl, retargeted to G1 with OmniRetarget, and augmented with HumanoidMimicGen. Residual RL on SONIC produces successful LeRobot demonstrations for GR00T initialization. Our DAgger loop restarts the privileged expert from a sampled state of a failed VLA rollout, follows the same reference suffix, and aggregates successful recoveries. The final VLA is evaluated on four simulated tasks. Motion and simulation reference images are adapted from DreamControl~\citep{dreamcontrol}.}
\label{fig:pipeline}
\end{figure}
}

\section{Introduction}
\label{sec:intro}

Vision-language-action (VLA) models offer a common interface for mapping visual observations and natural-language instructions to robot behavior. Models such as OpenVLA and GR00T learn from robot demonstrations and can be adapted to downstream tasks through additional training~\citep{openvla,groot}. For a humanoid, these demonstrations must coordinate reaching, hand interaction, torso motion, foot placement, and balance. A hierarchical interface can use a low-level (system-1) controller such as SONIC~\citep{sonic} to execute robot motion tokens supplied by a higher-level policy or a human teleoperation interface. Using this interface, an operator collects interactive demonstrations for VLA training. Collecting this behavior repeatedly across scene configurations demands operator time, a suitable teleoperation interface, and careful supervision. Increasing the number of tasks and environmental variations therefore increases the human effort required to build useful training data.

Simulation offers a way to scale this data, but replacing a teleoperator with an autonomous data generator introduces a control problem. Reinforcement learning in simulation with sparse task success gives a high-dimensional humanoid little guidance for approaching a target while maintaining balance \citep{dreamcontrol}. Dense rewards can support exploration, yet often require task-specific choices about reaching, posture, contact, timing, and object progress \citep{viral,doorman}. A policy can also satisfy parts of a reward while producing behavior that is poorly suited to imitation. Consequently, the usefulness of simulated demonstrations depends on both physical task completion and the quality of the behavior used to achieve it.

Generative human motion models provide a complementary source of structure. Text and spatial constraints can specify the intended behavior and the location of an interaction, yielding whole-body motion references with a human motion prior. DreamControl combines such references with RL for humanoid scene interaction~\citep{dreamcontrol}. This motivates using generated motion as a scalable source of supervision for downstream VLA training. The remaining challenge is to turn kinematic intent into repeatable interactions with movable objects, and then transfer that competence to a policy that receives images and language. Reward design remains part of this process, but references provide a unified framework for task-reward specification.

We propose \method{}, a framework for producing task demonstrations in simulation from generated human motions and distilling them into a VLA. We first generate task-conditioned references with DreamControl, using OmniControl to constrain relevant human joints at target locations and times~\citep{omnicontrol}. After retargeting these motions to the Unitree G1, we train an expert to co-track the humanoid and the interacting object. We use SONIC as the whole-body control backbone~\citep{sonic}, and learn residual corrections in its motion-token interface to adapt the reference behavior to contact and randomized physics. The resulting expert acts as an automated demonstration generator: motion synthesis supplies the intended human-like behavior, and RL supplies the physically executed interaction. Successful rollouts are recorded as a LeRobot dataset and used to fine-tune GR00T.

This construction is analogous to collecting teleoperation data with a generative model supplying the motion commands. It removes the need to have a human operate the robot for every downstream demonstration while retaining an interface suited to whole-body control. The human motion prior and the pretrained control and VLA backbones remain valuable upstream resources. Our claim concerns eliminating \emph{new task-specific teleoperation collection} for the simulated tasks, rather than eliminating human data from the pretraining of those resources. We formulate each interactive task using motion references, object behavior, and a success specification; the present study evaluates four tasks.

The central research question is how to distill this expert into a VLA that remains competent under its own closed-loop execution. Successful expert trajectories are a useful starting point, but they emphasize states reached by the expert. A student that grasps slightly late, approaches a handle at a different angle, or misaligns its fingertip can visit states absent from that dataset. Its next action changes the next observation, allowing small errors to compound. Collecting more clean demonstrations may improve coverage of nominal behavior without providing the corrective behavior needed after these deviations. Offline imitation alone therefore leaves a gap between supervised action prediction and reliable interaction.

We address this gap through a reference-conditioned DAgger distillation procedure that collects corrective trajectories after learner failures. Standard DAgger aggregates expert labels on states encountered by the learner~\citep{dagger}, and prior humanoid teacher--student systems such as VIRAL already use online DAgger~\citep{viral}. Our task-specific adaptation queries a generated-motion, residual-SONIC expert at a randomly sampled state of a failed learner rollout. The expert resumes with the suffix of the same generated reference, and only successful recovery trajectories are added to the aggregate. This supplies temporally consistent corrective demonstrations for the language-conditioned VLA while retaining the reference context required by the expert. 

The evaluation is organized around picking from a platform, opening a drawer, pressing a button, and walking and picking from a platform in simulation. These tasks expose distinct interaction requirements: establishing and maintaining a grasp, driving an articulated object, reaching a precise contact target, and combining locomotion with grasping. We present results from controlled studies of the number of DAgger iterations, the residual control formulation, and the amount of generated training data. Together, task completion, motion fidelity, and the student--expert performance gap characterize demonstration quality and the competence retained through distillation.

Our three contributions are:
\begin{enumerate}
    \item \textbf{A generative demonstration framework.} We propose a pipeline that uses generated human motions and residual RL on a SONIC backbone to collect natural, human-like interaction demonstrations for VLA training in simulation, without task-specific human teleoperation collection.
    \item \textbf{A DAgger-based VLA distillation method.} We propose using the RL expert to supervise the VLA on learner-visited states, with reference-conditioned expert recovery and success-filtered trajectory aggregation as the methodological contribution. 
    \item \textbf{A four-task simulation study with controlled ablations.} We evaluate platform picking, drawer opening, button pressing, and walking and picking with controlled ablations of DAgger iterations, residual RL, and generated data scale.
\end{enumerate}

\section{Related Work}
\label{sec:relatedwork}

\subsection{Humanoid Motion Tracking and Teleoperation}

Human-motion priors support humanoid tracking and whole-body control~\citep{deepmimic,amp,phc,exbody2,beyondmimic,asap,pgmt}, while camera, VR, and portable interfaces support teleoperation and demonstration collection~\citep{h2o,omnih2o,humanplus,twist,humanoidumi,wb_umi}. We use SONIC's motion-token interface~\citep{sonic} to turn generated references into autonomous interaction demonstrations. Generated motions supply the nominal commands in our pipeline, and task-specific residual RL adjusts their execution at contact without collecting new teleoperated demonstrations.

\subsection{Vision-Language-Action Learning for Humanoids}

Humanoid VLAs connect visual-language representations to executable behavior through multimodal pretraining and reusable motor controllers~\citep{groot,groot16,groot17,helix,helix02,helix25}. Synthetic kinematic and optimal-control demonstrations also provide supervision for whole-body visual policies~\citep{vlk,wolfvla}. DreamVLA uses a generated-motion, residual-SONIC expert to produce physically executed interactions and reference-conditioned corrective trajectories for GR00T. We fine-tune an existing VLA backbone and study demonstration generation and corrective supervision, without introducing a new model architecture.

\subsection{Visual Sim-to-Real Across Embodiments}

Across manipulation, legged, and aerial embodiments, visual sim-to-real combines simulated experience with perception randomization, privileged supervision, or reconstructed scenes~\citep{domainrandomization,dexhand,dextreme,viserdex,rialto,transic,dextera,molmobot,flare,perceptivelocomotion,parkour,umi_on_legs,navila,flight}. Humanoid pipelines use teacher--student distillation~\citep{viral,doorman}. We study VLA distillation in simulation. Our pipeline shares privileged-to-visual supervision with these humanoid systems, while isolating motion-guided expert training and corrective VLA distillation without visual domain randomization.

\subsection{Motion Generation, Retargeting, and Synthetic Demonstrations}

Language-conditioned motion models provide interaction references~\citep{mdm,t2mgpt,motiongpt,omnicontrol,omomo,chois}, while retargeting and demonstration augmentation adapt motions to robot morphology and new scenes~\citep{gmr,omniretarget,mimicgen,skillgen,humanoidmimicgen}. Generative video and simulated experts offer complementary routes to synthetic policy data~\citep{gen2act,dreamgen,mimicdreamer,dreamzero,video2policy}. We build on DreamControl~\citep{dreamcontrol} to produce physically executed SONIC demonstrations from guided human motion. Residual RL resolves the contact dynamics left by kinematic synthesis, and successful-rollout filtering converts the resulting interactions into executable action labels for VLA training.

\subsection{Expert Distillation and Corrective Data Aggregation}

DAgger-style distillation addresses learner-induced distribution shift~\citep{dagger,hgdagger,rewarddagger,viral}. Our adaptation restores a state from a failed VLA rollout, resumes the expert with the same reference suffix, and aggregates only successful corrective trajectories for GR00T action-chunk training. This retains the DAgger principle of querying learner-visited states while collecting full, phase-consistent recovery segments for action-chunk training.

\section{Methodology}
\label{sec:method}

\method{} comprises four stages: reference generation, residual expert training and successful rollout collection, GR00T distillation with iterative expert recovery, and simulation validation (Figure~\ref{fig:pipeline}). Let $\ell$ denote a task instruction, $s_t$ the full simulator state, and $o_t=(I_t,p_t)$ the student observation, consisting of camera images $I_t$ and robot proprioception $p_t$. The expert uses privileged simulator information; the VLA predicts executable commands from $o_t$ and $\ell$. Privileged object states used for expert training and success checks are excluded from the student observation.

\subsection{Generating Task References with DreamControl}
\label{sec:generation}

We follow DreamControl's trajectory-generation stage~\citep{dreamcontrol}. OmniControl~\citep{omnicontrol} samples human motions conditioned on an instruction and spatiotemporal constraints,
\begin{equation}
    H \sim p_{\mathrm{motion}}(H\mid\ell,\mathcal{G}),
    \qquad \mathcal{G}=\{(j_k,\mathbf{v}_k,t_k)\}_{k=1}^{K},
    \label{eq:motion}
\end{equation}
where joint $j_k$ is guided toward position $\mathbf{v}_k$ at time $t_k$. Sampling target locations and interaction times produces varied references. Wrist targets specify the picking or drawer interaction; an appropriate hand target specifies button contact.

The generated motions are retargeted to G1 using OmniRetarget~\citep{omniretarget}. After retargeting, we have a set of reference kinematic trajectories denoted by $\mathcal{K} = \{\mathcal{k}_0,\mathcal{k}_1,\ldots,\mathcal{k}_{N-1}\}$.
We represent each robot reference by root pose, body joint configuration, and hand state, and align the simulated scene with its intended interaction frame. We also augment the generated trajectories by modifying the contact points using HumanoidMimicGen~\citep{humanoidmimicgen}.

\subsection{Residual RL for Humanoid--Object Co-Tracking}
\label{sec:rl}

\paragraph{Expert inputs and object references.}
The privileged expert observes robot joint positions and velocities, root linear and angular velocities, gravity projected into the root frame, the previous executed command, and task-relevant object state $x_t$. Following DreamControl's privileged observation representation~\citep{dreamcontrol}, we encode the local reference segment $R_{t:t+F}$ as current and look-ahead target observations:
\begin{equation}
    \mathcal{O}_t^{\mathrm{ref}} = [\gamma_{t\mid t},\gamma_{t+\Delta t_{\mathrm{obs}}\mid t},
        \ldots,\gamma_{t+F\mid t}],
    \label{eq:referencewindow}
\end{equation}
where $\Delta t_{\mathrm{obs}}=0.1\,\mathrm{s}$, $K_{\mathrm{ref}}$ is the number of sampled frames, and $F=(K_{\mathrm{ref}}-1)\Delta t_{\mathrm{obs}}$. For reference time $\tau$, each target observation is
\begin{equation}
    \gamma_{\tau\mid t} = \bigl(q_{\tau}^{\mathrm{ref}},\dot{q}_{\tau}^{\mathrm{ref}},
        p_{\tau\mid t}^{\mathrm{rel,root}},p_{\tau\mid t}^{\mathrm{rel,key}},
        s_{\tau}^{\mathrm{ref,left}},s_{\tau}^{\mathrm{ref,right}}\bigr).
    \label{eq:referenceframe}
\end{equation}
Here, $q_{\tau}^{\mathrm{ref}},\dot{q}_{\tau}^{\mathrm{ref}}\in\mathbb{R}^{27}$ specify target joint angles and velocities; $p_{\tau\mid t}^{\mathrm{rel,root}}$ represents the reference root's position and orientation relative to the robot's current base; and $p_{\tau\mid t}^{\mathrm{rel,key}}\in\mathbb{R}^{3\times41}$ contains reference keypoint coordinates in that same frame. The hand targets $s_{\tau}^{\mathrm{ref,left}},s_{\tau}^{\mathrm{ref,right}}\in\{0,1\}$ encode open ($0$) and closed ($1$) configurations. For keypoint $j$,
\begin{equation}
    p_{\tau\mid t}^{\mathrm{rel,key},j}
        = Q_t^{\mathsf{T}}\bigl(p_{\tau}^{\mathrm{ref,key},j}
            -p_t^{\mathrm{robot,root}}\bigr),
    \label{eq:referencecoordinates}
\end{equation}
where $Q_t$ is the robot root's rotation in world coordinates. Using the current robot frame for all targets preserves spatial tracking errors relevant to contact.

The object reference $\bar{x}_t$ separately describes intended interaction progress: object pose for picking, drawer joint displacement for opening, or button travel for pressing. Co-tracking couples the humanoid reference to the corresponding object outcome, so matching a hand trajectory alone does not satisfy the interaction objective. Reference windows and privileged object state are used by the RL expert; the VLA executes without these inputs.

\paragraph{SONIC backbone and residual token actions.}
SONIC maps motion commands into a shared token representation and decodes them with proprioception into robot control commands~\citep{sonic}. We use this interface as the backbone for a residual task expert. Let $E_{\mathrm{S}}$ denote the motion encoder and $D_{\mathrm{S}}$ the control decoder. We write the residual formulation as
\begin{align}
    z_t^{\mathrm{nom}} &= E_{\mathrm{S}}(R_{t:t+F}), &
    \Delta z_t &= \pi_{\mathrm{res}}(s_t,R_{t:t+F},\bar{x}_{t:t+F},\ell),\\
    z_t^{\mathrm{E}} &= z_t^{\mathrm{nom}} + \Delta z_t, &
    a_t^{\mathrm{body}} &= D_{\mathrm{S}}(p_t,z_t^{\mathrm{E}}).
    \label{eq:residual}
\end{align}
The residual adapts the generated motion to the actual robot--object interaction, while SONIC provides the underlying whole-body control prior. The hand commands $a_t^\text{hand}$ complete the executable expert command $u_t^{\mathrm{E}}= \{z_t^{\mathrm{E}},a_t^\text{hand}\}$. We operate the hand in a binary open/close mode i.e. the hand action $a_t^\text{hand}$ is binary. 

\paragraph{Training objective.}
The expert is trained to maximize return under a reward comprising humanoid reference tracking, sparse task-specific rewards, and control regularization:
\begin{equation}
    r_t = w_{\mathrm{hum}}r_{\mathrm{hum}}(s_t,R_t)
        + w_{\mathrm{task}}r_{\mathrm{task}}(s_t)
        - w_{\mathrm{reg}}c_{\mathrm{reg}}(u_t^{\mathrm{E}},u_{t-1}^{\mathrm{E}}).
    \label{eq:reward}
\end{equation}
The reward structure for the tasks is the same as in DreamControl \citep{dreamcontrol}. The tracking objectives supply dense reference supervision; the task objective checks meaningful interaction progress or completion. The regularization term discourages excessive corrections and abrupt commands. Reference tracking guides the motion strategy, while sparse task rewards specify the interaction outcome.

\paragraph{Physics domain randomization.}
We randomize robot and object physics during expert training to expose the policy to variation in interaction dynamics. Randomized scenes preserve feasibility and remain compatible with their target references. Physics randomization addresses control robustness; the VLA is trained without visual domain randomization.

\subsection{Successful Expert Rollouts as a LeRobot Dataset}
\label{sec:data}

After expert training, we execute rollouts over sampled scenes, references, and physical parameters. An episode is accepted only when its task-specific success predicate $\mathsf{S}(\tau)$ is satisfied. We export synchronized images, proprioceptive states, actions, timestamps, episode identifiers, and task instructions in LeRobot format~\citep{lerobot}. The initial imitation dataset is
\begin{equation}
    \dataset_0 = \bigcup_{\tau:\,\mathsf{S}(\tau)=1}
        \{(o_t,\ell,u_t^{\mathrm{E}})\}_{t=0}^{T_{\tau}-1}.
    \label{eq:successdata}
\end{equation}
where $u_t^{\mathrm{E}}= \{z_t^{\mathrm{E}},a_t^\text{hand}\}$ represents the composed SONIC motion command and the hand command, $\ell$ is the language prompt, $o_t = \{I_t,p_t\}$ is the observation consisting of camera images, $I_t$, and proprioception $p_t$. We use only the robot's egocentric camera to obtain $I_t$. By predicting these executable commands, the VLA uses the same low-level backbone without a reference trajectory at inference.
Each accepted episode contains the executed interaction, including the approach and completion phases. This produces successful demonstrations suitable for VLA supervision from physically simulated expert behavior. 

Success filtering improves the quality of the initial demonstration set, but changes its distribution. Difficult scenes may contribute to fewer accepted episodes and nominal successes may contain little corrective behavior. We therefore track both the number of attempted episodes and the number retained, along with scene coverage. The DAgger stage extends this dataset with successful expert recoveries from learner-induced failures.

\subsection{GR00T Fine-Tuning and DAgger-Based Distillation}
\label{sec:dagger}

\paragraph{Offline initialization.}
We fine-tune a pretrained GR00T VLA~\citep{groot} on $\dataset_0$, conditioning action prediction on visual observations, robot state, and task instruction. Let $U_t^{\mathrm{E}}=(u_t^{\mathrm{E}},\ldots,u_{t+h-1}^{\mathrm{E}})$ denote an expert action chunk of horizon $h$. Constructing action chunks from consecutive samples within each episode, the initialization objective is
\begin{equation}
    \theta_0 = \arg\min_{\theta}
    \mathbb{E}_{(o_t,\ell,U_t^{\mathrm{E}})\sim\dataset_0}
    \left[\mathcal{L}_{\mathrm{GR00T}}(\theta;o_t,\ell,U_t^{\mathrm{E}})\right].
    \label{eq:bc1}
\end{equation}
The loss is the native action-training objective of the selected GR00T implementation. Privileged object state and reference context remain available to the expert but are excluded from the student.

\paragraph{Why iterative supervision is needed.}
The initialized student is trained on expert-generated observations and then evaluated on observations induced by its own actions. These distributions differ whenever perception, timing, or action prediction changes the interaction. This matters particularly at contact: a small error can alter whether an object is grasped, whether a drawer handle is engaged, or whether a button is actuated. The appropriate next command may then differ substantially from the next command in the original successful trajectory. Iterative distillation exposes the VLA to such states and obtains expert supervision appropriate to the current interaction.

\paragraph{Standard DAgger foundation.}
At iteration $k$, the current student is rolled out and the expert labels selected visited states. Denoting that set of states by $\mathcal{V}_k$, the standard aggregation structure is
\begin{equation}
    \dataset_{k+1} = \dataset_k \cup
    \{(o_t,\ell,\expert(s_t,c_t)):\ s_t\in\mathcal{V}_k\},
    \label{eq:dagger}
\end{equation}
followed by training the next student on the aggregate~\citep{dagger}. The context $c_t$ denotes whatever reference and task information is required to make an expert query well defined. Equation~\ref{eq:dagger} describes the established DAgger principle. 

\paragraph{Reference-conditioned recovery DAgger.}
Our adaptation collects successful expert recovery segments seeded by states from failed learner rollouts, rather than independently labeling every learner state. At iteration $k$, we use the following procedure:
\begin{enumerate}
    \item Sample a reference $\mathcal{k}_i\in\mathcal{K}$ with length $T_i$ and initialize the robot and object according to that reference.
    \item Roll out the current VLA $\pi_{\theta_k}$ until task success or timeout at $T_i$, saving simulator states along the learner trajectory. Successful learner rollouts require no corrective trajectory in this procedure.
    \item For a failed rollout, sample $t$ uniformly from $\{0,\ldots,T_i-1\}$ and restore the saved learner state $s_t$. Resume the privileged RL expert using the same reference $\mathcal{k}_i$ starting at phase $t$. Retain the resulting corrective segment only if the expert completes the task within the remaining reference horizon; otherwise discard it.
\end{enumerate}
Let $\widetilde{\tau}_{i,t}$ denote this expert recovery segment, and let $\widetilde{s}_j$, $\widetilde{o}_j$, and $\widetilde{u}_j^{\mathrm{E}}$ denote its states, observations, and executed expert commands. The initial recovery state satisfies $\widetilde{s}_t=s_t$, while subsequent states are induced by the expert. Aggregation therefore takes the trajectory-level form
\begin{equation}
    \dataset_{k+1}=\dataset_k\cup
    \bigcup_{\widetilde{\tau}_{i,t}:\,\mathsf{S}(\widetilde{\tau}_{i,t})=1}
    \{(\widetilde{o}_j,\ell,\widetilde{u}_j^{\mathrm{E}})\}_{j=t}^{T^{\mathrm{rec}}_{i,t}-1},
    \label{eq:recoverydagger}
\end{equation}
where $T^{\mathrm{rec}}_{i,t}\leq T_i$ is the recovery termination time in the original reference clock. We fine-tune GR00T on the aggregate and repeat. Action chunks are constructed within each accepted segment, without crossing episode or recovery boundaries. The student uses the same visual, proprioceptive, and language inputs as during offline initialization; simulator state and the reference suffix are used only to query the expert. This makes the distinction from standard pointwise DAgger explicit: supervision is provided by successful, reference-conditioned expert continuations from learner-induced failures.

\section{Results}
\label{sec:results}

We evaluate \method{} on four simulated interaction tasks and study the effects of DAgger iterations, residual control, and generated dataset scale.

\subsection{Four-Task Setup and the Effect of DAgger Iterations}
\label{sec:setup}

\begin{figure}[t]
\centering
\includegraphics[width=\linewidth]{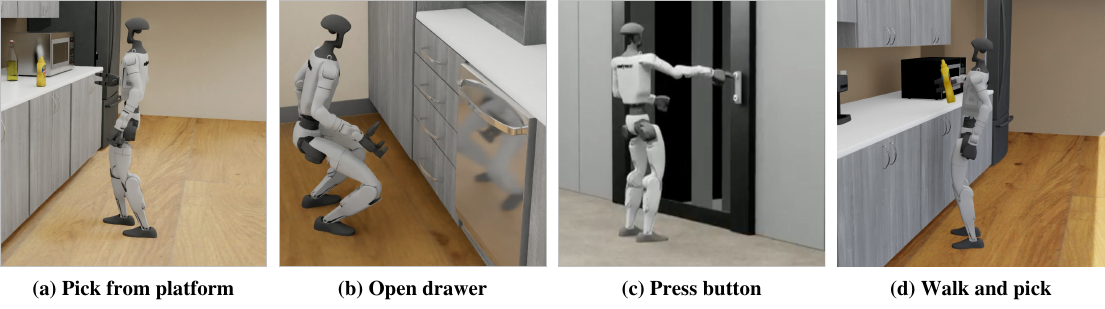}
\caption{Four-task simulation setup with the Unitree G1: (a) picking from a platform, (b) opening a drawer, (c) pressing a button, and (d) walking and picking from a platform. The screenshots are cropped to equal-sized panels with comparable robot heights.}
\label{fig:tasks}
\end{figure}

\paragraph{Task suite.}
We evaluate a simulated Unitree G1 on the four interactive tasks in Table~\ref{tab:tasks}. Each task requires whole-body coordination and a verifiable task outcome. Table~\ref{tab:tasks} defines the task-specific success criteria. Figure~\ref{fig:tasks} shows representative simulation scenes for the four tasks.

\begin{table}[t]
\caption{Task setup and success criteria for the four simulated interactions.}
\label{tab:tasks}
\centering
\small
\begin{tabularx}{\linewidth}{@{}>{\raggedright\arraybackslash}p{0.85in}YY@{}}
\toprule
Task & Interaction setup & Success predicate \\
\midrule
Pick from platform & Reach an object on a platform, grasp it, and lift while maintaining balance. & Object is lifted at least 15\,cm above the platform and remains held. \\
Open drawer & Approach the handle, establish contact or grasp, and pull along the drawer's articulation. & Drawer displacement exceeds 0.15\,m. \\
Press button & Align the interacting hand with the button and establish actuating contact. & The robot's right index finger contacts the button. \\
Walk and pick & Walk to a platform with an object on it, reach for the object, grasp it, and lift while maintaining balance. & Object is lifted at least 15 cm above the platform and remains held.\\
\bottomrule
\end{tabularx}
\end{table}

\paragraph{Evaluation and reporting.}
The primary metric is task success rate. The distribution of object, button, or drawer-handle locations is the same as the distribution from which target positions are chosen for generating their trajectories. We also report per-task expert success, student performance relative to the expert, and mean per-joint position error (MPJPE). For each task-success rate evaluation, we use $100$ rollouts.

\paragraph{DAgger iteration ablation.}
We compare the offline student ($k=0$) with students after $1$, $2$, and $3$ aggregation rounds (Table~\ref{tab:dagger}). We use $200$ initial expert trajectories and collect $200$ additional successful corrective trajectories per round, giving $800$ trajectories at $k=3$. The matched offline-data control uses $800$ successful trajectories collected from the base expert without learner-triggered recovery, matching the total trajectory budget of three DAgger rounds. This controls for the amount of data when evaluating the usefulness of corrective aggregation.

\begin{table}[t]
\caption{DAgger results on all tasks. Success is reported per task with $100$ random environment settings.}
\label{tab:dagger}
\centering
\small
\begin{tabularx}{\linewidth}{@{}Ycccccc@{}}
\toprule
Distillation variant & Rounds & Trajectories & Pick & Drawer & Button & \begin{tabular}[t]{@{}c@{}}Walk and\\Pick\end{tabular} \\
\midrule
Offline GR00T initialization & $0$ & $200$ & $0\%$ & $51\%$ & $59\%$ & $0\%$ \\
DAgger with $k=1$ & $1$ & $400$ & $17\%$ & $55\%$ & $58\%$ & $1\%$ \\
DAgger with $k=2$ & $2$ & $600$ & $30\%$ & $54\%$ & $62\%$ & $7\%$ \\
DAgger with $k=3$ & $3$ & $800$ & $54\%$ & $60\%$ & $69\%$ & $18\%$ \\
Matched offline-data control & $0$ & $800$ & $0\%$ & $54\%$ & $57\%$ & $0\%$ \\
\bottomrule
\end{tabularx}
\end{table}

\subsection{Effectiveness of Residual RL on SONIC}
\label{sec:residualresults}

\begin{figure}[t]
\centering
\includegraphics[width=.9\linewidth]{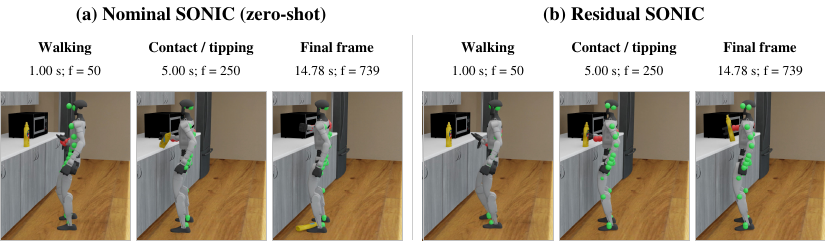}
\caption{Matched nominal and residual SONIC execution on the mustard-bottle interaction. Both subfigures use video frames $50$, $250$, and $739$ (the final frame), corresponding to $1.00$, $5.00$, and $14.78$\,s, with identical crop coordinates. (a) Left: nominal zero-shot tracking tips the bottle off the counter; the bottle is on the floor at the end. (b) Right: residual control retains the bottle in the hand.}
\label{fig:residual}
\end{figure}

This study isolates the choice to learn residual SONIC token commands. Table~\ref{tab:residual} compares expert task success and tracking error after the same budget of $3000$ RL training iterations for the learned formulations. Residual token RL achieves higher task success than absolute token RL under this matched training budget. We also report performance of using SONIC commands as-is with no residual RL. Without residual adaptation, nominal SONIC tracking has zero success on picking and lower success than residual RL on the other tasks (Table~\ref{tab:residual}). Tracking error in the base SONIC controller and kinematic references that are not exactly trackable can hinder zero-shot execution on precise tasks. Residual RL makes small trajectory corrections needed to complete the task while retaining the nominal SONIC motion, and compensates for the forces needed to pick an object or open a heavy drawer. Figure~\ref{fig:residual} compares matched video frames from nominal and residual execution. In this rollout, nominal SONIC tips the mustard bottle off the counter, whereas residual control retains the bottle in the hand.

\begin{table}[t]
\caption{Expert task success and tracking error for nominal SONIC tracking, residual token RL, and absolute token RL. MPJPE is mean per-joint position error; lower is better.}
\label{tab:residual}
\centering
\small
\begin{tabularx}{\linewidth}{@{}Yccccc@{}}
\toprule
Expert formulation & Pick & Drawer & Button & \begin{tabular}[t]{@{}c@{}}Walk and\\Pick\end{tabular} & \begin{tabular}[t]{@{}c@{}}MPJPE\\(mm)\end{tabular} \\
\midrule
Nominal SONIC tracking & $0\%$ & $30.0\%$ & $68.0\%$ & $17.30\%$ & $24.84$ \\
Residual SONIC token RL & $88.6\%$ & $74.0\%$ & $81.0\%$ & $75.95\%$ & $25.51$ \\
Absolute token RL control & $0\%$ & $0\%$ & $0\%$ & $0\%$ & $62.29$ \\
\bottomrule
\end{tabularx}
\end{table}

\subsection{Scaling the Generated Dataset}
\label{sec:scaling}

We vary the number of generated kinematic references while keeping the task mix, observation/action schema, and evaluation set fixed. Nested subsets sampled at the episode level allow comparisons across data budgets. Increasing the number of generated references leads to more diverse behavior and coverage, and higher success rates for drawer opening and button pressing; picking and walking and picking remain at zero success in this ablation. Table~\ref{tab:scaling} reports episode counts and per-task success across reference budgets.

\begin{table}[!htbp]
\caption{Generated-data scaling. Nested reference budgets are fixed at $10$, $50$, and $200$; episode counts and held-out task success are reported for each budget.}
\label{tab:scaling}
\centering
\small
\begin{tabularx}{\linewidth}{@{}Yccccc@{}}
\toprule
References & Episodes & Pick & Drawer & Button & \begin{tabular}[t]{@{}c@{}}Walk and\\Pick\end{tabular} \\
\midrule
10 & $10$ & $0\%$ & $13.0\%$ & $0\%$ & $0\%$ \\
50 & $50$ & $0\%$ & $51.0\%$ & $14.0\%$ & $0\%$ \\
200 & $200$ & $0\%$ & $51.0\%$ & $59.0\%$ & $0\%$ \\
\bottomrule
\end{tabularx}
\end{table}

\section{Limitations}
\label{sec:limitations}

The evaluation is limited to four simulated interaction tasks on a single humanoid embodiment. Generated motions can be incompatible with robot kinematics or contact requirements, and task-specific guidance, filtering, object references, and reward weights remain necessary parts of the pipeline. Successful-rollout filtering may underrepresent difficult configurations. Reference-conditioned experts may provide unreliable or abrupt corrections on learner-visited states, particularly during precise interactions.

\section{Conclusion and Future Work}
\label{sec:conclusion}

We present a framework for turning generated human motion into simulated humanoid demonstrations and using them to train a VLA. DreamControl supplies task-conditioned references, residual RL on a SONIC backbone adapts them to robot--object interaction, and successful rollouts initialize GR00T fine-tuning. Reference-conditioned recovery DAgger distills expert behavior under learner-induced state distributions by aggregating successful corrective continuations.

Future work focuses on expanding the task set and the diversity of interacting objects, studying longer task sequences, and quantifying data-generation efficiency at larger scale. A further direction is sim-to-real deployment, informed by the visual humanoid transfer pipelines of DoorMan and VIRAL~\citep{doorman,viral}. Key requirements include validating the control and sensing stack on hardware, matching cameras and contact behavior, and studying visual and physical randomization. These directions extend the present simulation study; hardware deployment is outside its scope.

\enlargethispage{2\baselineskip}
\acknowledgments{This research was supported in part by DOD Advanced Research Projects Agency project Design of Robustly Implementable Autonomous and Intelligent Machines under Award HR00112490425.}


\clearpage
\begin{thebibliography}{99}

\bibitem[Kim et~al.(2024)]{openvla}
M.~J. Kim, K.~Pertsch, S.~Karamcheti, et~al.
\newblock OpenVLA: An Open-Source Vision-Language-Action Model.
\newblock \emph{arXiv:2406.09246}, 2024.
\newblock \url{https://arxiv.org/abs/2406.09246}.

\bibitem[Bjorck et~al.(2025)]{groot}
J.~Bjorck, F.~Casta\~neda, N.~Cherniadev, et~al.
\newblock GR00T N1: An Open Foundation Model for Generalist Humanoid Robots.
\newblock \emph{arXiv:2503.14734}, 2025.
\newblock \url{https://arxiv.org/abs/2503.14734}.

\bibitem[Kalaria et~al.(2025)]{dreamcontrol}
D.~Kalaria, S.~Harithas, P.~Katara, S.~Kwak, S.~Bhagat, S.~S. Sastry,
S.~Sridhar, S.~Vemprala, A.~Kapoor, and J.~Huang.
\newblock DreamControl: Human-Inspired Whole-Body Humanoid Control for Scene Interaction via Guided Diffusion.
\newblock \emph{arXiv:2509.14353}, 2025.
\newblock \url{https://genrobo.github.io/DreamControl/static/paper/DreamControl_arxiv.pdf}.

\bibitem[Xie et~al.(2024)]{omnicontrol}
Y.~Xie, V.~Jampani, L.~Zhong, D.~Sun, and H.~Jiang.
\newblock OmniControl: Control Any Joint at Any Time for Human Motion Generation.
\newblock In \emph{International Conference on Learning Representations}, 2024.
\newblock \url{https://arxiv.org/abs/2310.08580}.

\bibitem[Yang et~al.(2025)]{omniretarget}
L.~Yang, X.~Huang, Z.~Wu, A.~Kanazawa, P.~Abbeel, C.~Sferrazza,
C.~K. Liu, R.~Duan, and G.~Shi.
\newblock OmniRetarget: Interaction-Preserving Data Generation for Humanoid Whole-Body Loco-Manipulation and Scene Interaction.
\newblock \emph{arXiv:2509.26633}, 2025.
\newblock \url{https://arxiv.org/abs/2509.26633}.

\bibitem[Lin et~al.(2026)]{humanoidmimicgen}
K.~Lin, A.~Mandlekar, C.~R. Garrett, N.~Chernyadev, Y.~Fang,
R.~Ding, Y.~Xie, J.~Tran, L.~Fan, and Y.~Zhu.
\newblock HumanoidMimicGen: Data Generation for Loco-Manipulation via Whole-Body Planning.
\newblock \emph{arXiv:2605.27724}, 2026.
\newblock \url{https://arxiv.org/abs/2605.27724}.

\bibitem[Luo et~al.(2026)]{sonic}
Z.~Luo, Y.~Yuan, T.~Wang, C.~Li, F.~Casta\~neda, et~al.
\newblock SONIC: Supersizing Motion Tracking for Natural Humanoid Whole-Body Control.
\newblock \emph{Science Robotics}, 11(117):eaed4592, 2026.
\newblock \doi{10.1126/scirobotics.aed4592}.
\newblock \url{https://arxiv.org/abs/2511.07820}.

\bibitem[Ross et~al.(2011)]{dagger}
S.~Ross, G.~Gordon, and D.~Bagnell.
\newblock A Reduction of Imitation Learning and Structured Prediction to No-Regret Online Learning.
\newblock In \emph{Proceedings of the Fourteenth International Conference on Artificial Intelligence and Statistics}, PMLR 15:627--635, 2011.
\newblock \url{https://proceedings.mlr.press/v15/ross11a.html}.

\bibitem[He et~al.(2025)]{viral}
T.~He, Z.~Wang, H.~Xue, Q.~Ben, Z.~Luo, W.~Xiao, Y.~Yuan, X.~Da,
F.~Casta\~neda, S.~Sastry, C.~Liu, G.~Shi, L.~Fan, and Y.~Zhu.
\newblock VIRAL: Visual Sim-to-Real at Scale for Humanoid Loco-Manipulation.
\newblock \emph{arXiv:2511.15200}, 2025.
\newblock \url{https://arxiv.org/abs/2511.15200}.

\bibitem[Hugging Face(2026)]{lerobot}
Hugging Face.
\newblock LeRobotDataset v3.0: Dataset documentation.
\newblock Online documentation, accessed October 2, 2026.
\newblock \url{https://huggingface.co/docs/lerobot/lerobot-dataset-v3}.

\bibitem[NVIDIA(2026)]{grootdata}
NVIDIA.
\newblock Isaac-GR00T: Robot Data Preparation Guide.
\newblock Online documentation, accessed October 2, 2026.
\newblock \url{https://github.com/NVIDIA/Isaac-GR00T/blob/main/getting_started/data_preparation.md}.

\bibitem[Xue et~al.(2025)]{doorman}
H.~Xue, T.~He, Z.~Wang, Q.~Ben, W.~Xiao, Z.~Luo, X.~Da,
F.~Casta\~neda, G.~Shi, S.~Sastry, L.~Fan, and Y.~Zhu.
\newblock Opening the Sim-to-Real Door for Humanoid Pixel-to-Action Policy Transfer.
\newblock \emph{arXiv:2512.01061}, 2025.
\newblock \url{https://arxiv.org/abs/2512.01061}.

\bibitem[Peng et~al.(2018)]{deepmimic}
X.~B.~Peng, P.~Abbeel, S.~Levine, et~al.
\newblock DeepMimic: Example-Guided Deep Reinforcement Learning of Physics-Based Character Skills.
\newblock \emph{arXiv:1804.02717}, 2018.
\newblock \url{https://arxiv.org/abs/1804.02717}.

\bibitem[Peng et~al.(2021)]{amp}
X.~B.~Peng, Z.~Ma, P.~Abbeel, et~al.
\newblock AMP: Adversarial Motion Priors for Stylized Physics-Based Character Control.
\newblock \emph{arXiv:2104.02180}, 2021.
\newblock \url{https://arxiv.org/abs/2104.02180}.

\bibitem[Luo et~al.(2023)]{phc}
Z.~Luo, J.~Cao, A.~Winkler, et~al.
\newblock Perpetual Humanoid Control for Real-time Simulated Avatars.
\newblock \emph{arXiv:2305.06456}, 2023.
\newblock \url{https://arxiv.org/abs/2305.06456}.

\bibitem[Ji et~al.(2024)]{exbody2}
M.~Ji, X.~Peng, F.~Liu, et~al.
\newblock ExBody2: Advanced Expressive Humanoid Whole-Body Control.
\newblock \emph{arXiv:2412.13196}, 2024.
\newblock \url{https://arxiv.org/abs/2412.13196}.

\bibitem[Liao et~al.(2025)]{beyondmimic}
Q.~Liao, T.~E.~Truong, X.~Huang, et~al.
\newblock BeyondMimic: From Motion Tracking to Versatile Humanoid Control via Guided Diffusion.
\newblock \emph{arXiv:2508.08241}, 2025.
\newblock \url{https://arxiv.org/abs/2508.08241}.

\bibitem[He et~al.(2025)]{asap}
T.~He, J.~Gao, W.~Xiao, et~al.
\newblock ASAP: Aligning Simulation and Real-World Physics for Learning Agile Humanoid Whole-Body Skills.
\newblock \emph{arXiv:2502.01143}, 2025.
\newblock \url{https://arxiv.org/abs/2502.01143}.

\bibitem[Li et~al.(2026)]{pgmt}
H.~Li, P.~Li, Y.~Tao, et~al.
\newblock PGMT: Perceptive General Motion Tracking for Humanoid Robots.
\newblock \emph{arXiv:2609.08511}, 2026.
\newblock \url{https://arxiv.org/abs/2609.08511}.

\bibitem[He et~al.(2024)]{h2o}
T.~He, Z.~Luo, W.~Xiao, et~al.
\newblock Learning Human-to-Humanoid Real-Time Whole-Body Teleoperation.
\newblock \emph{arXiv:2403.04436}, 2024.
\newblock \url{https://arxiv.org/abs/2403.04436}.

\bibitem[He et~al.(2024)]{omnih2o}
T.~He, Z.~Luo, X.~He, et~al.
\newblock OmniH2O: Universal and Dexterous Human-to-Humanoid Whole-Body Teleoperation and Learning.
\newblock \emph{arXiv:2406.08858}, 2024.
\newblock \url{https://arxiv.org/abs/2406.08858}.

\bibitem[Fu et~al.(2024)]{humanplus}
Z.~Fu, Q.~Zhao, Q.~Wu, et~al.
\newblock HumanPlus: Humanoid Shadowing and Imitation from Humans.
\newblock \emph{arXiv:2406.10454}, 2024.
\newblock \url{https://arxiv.org/abs/2406.10454}.

\bibitem[Ze et~al.(2025)]{twist}
Y.~Ze, Z.~Chen, J.~P.~Ara{\'{u}}jo, et~al.
\newblock TWIST: Teleoperated Whole-Body Imitation System.
\newblock \emph{arXiv:2505.02833}, 2025.
\newblock \url{https://arxiv.org/abs/2505.02833}.

\bibitem[Wang et~al.(2026)]{humanoidumi}
H.~Wang, C.~Yu, Y.~Hu, et~al.
\newblock HumanoidUMI: Bridging Robot-Free Demonstrations and Humanoid Whole-Body Manipulation.
\newblock \emph{arXiv:2606.27239}, 2026.
\newblock \url{https://arxiv.org/abs/2606.27239}.

\bibitem[Nai et~al.(2026)]{wb_umi}
Y.~Nai, L.~Chang, L.~Yang, et~al.
\newblock Whole-Body UMI: Transferring UMI Manipulation Skills to Humanoid Whole-Body Manipulation via Real-Time Motion Generation.
\newblock \emph{arXiv:2609.22829}, 2026.
\newblock \url{https://arxiv.org/abs/2609.22829}.

\bibitem[NVIDIA(2025)]{groot16}
NVIDIA.
\newblock GR00T N1.6.
\newblock Model release and technical overview, 2025. Accessed October 8, 2026.
\newblock \url{https://research.nvidia.com/labs/gear/gr00t-n1_6/}.

\bibitem[NVIDIA(2026)]{groot17}
NVIDIA.
\newblock Isaac GR00T N1.7: Model Card.
\newblock Model card, 2026. Accessed October 8, 2026.
\newblock \url{https://huggingface.co/nvidia/GR00T-N1.7-3B}.

\bibitem[Figure(2025)]{helix}
Figure.
\newblock Helix: A Vision-Language-Action Model for Generalist Humanoid Control.
\newblock Technical report, 2025. Accessed October 8, 2026.
\newblock \url{https://www.figure.ai/news/helix}.

\bibitem[Figure(2026)]{helix02}
Figure.
\newblock Introducing Helix 02: Full-Body Autonomy.
\newblock Technical report, 2026. Accessed October 8, 2026.
\newblock \url{https://www.figure.ai/news/helix-02}.

\bibitem[Figure(2026)]{helix25}
Figure.
\newblock Helix 2.5: Zero-Shot 30-Home Generalization.
\newblock Technical report, 2026. Accessed October 8, 2026.
\newblock \url{https://www.figure.ai/news/helix-2-5-zero-shot-30-home-generalization}.

\bibitem[Wang et~al.(2026)]{vlk}
Y.~Wang, J.~Li, S.~Chen, et~al.
\newblock VLK: Learning Humanoid Loco-Manipulation from Synthetic Interactions in Reconstructed Scenes.
\newblock \emph{arXiv:2606.30645}, 2026.
\newblock \url{https://arxiv.org/abs/2606.30645}.

\bibitem[Boukheddimi et~al.(2026)]{wolfvla}
M.~Boukheddimi, O.~Adjali, D.~Sontag, et~al.
\newblock WOLF-VLA: Whole-Body Humanoid Optimal Locomotion Framework for Vision-Language-Action Learning.
\newblock \emph{arXiv:2606.25591}, 2026.
\newblock \url{https://arxiv.org/abs/2606.25591}.

\bibitem[Tobin et~al.(2017)]{domainrandomization}
J.~Tobin, R.~Fong, A.~Ray, et~al.
\newblock Domain Randomization for Transferring Deep Neural Networks from Simulation to the Real World.
\newblock \emph{arXiv:1703.06907}, 2017.
\newblock \url{https://arxiv.org/abs/1703.06907}.

\bibitem[OpenAI et~al.(2018)]{dexhand}
OpenAI, M.~Andrychowicz, B.~Baker, et~al.
\newblock Learning Dexterous In-Hand Manipulation.
\newblock \emph{arXiv:1808.00177}, 2018.
\newblock \url{https://arxiv.org/abs/1808.00177}.

\bibitem[Handa et~al.(2022)]{dextreme}
A.~Handa, A.~Allshire, V.~Makoviychuk, et~al.
\newblock DeXtreme: Transfer of Agile In-hand Manipulation from Simulation to Reality.
\newblock \emph{arXiv:2210.13702}, 2022.
\newblock \url{https://arxiv.org/abs/2210.13702}.

\bibitem[Bhardwaj et~al.(2026)]{viserdex}
A.~Bhardwaj, M.~Wilder-Smith, M.~Mittal, et~al.
\newblock ViserDex: Visual Sim-to-Real for Robust Dexterous In-hand Reorientation.
\newblock \emph{arXiv:2604.11138}, 2026.
\newblock \url{https://arxiv.org/abs/2604.11138}.

\bibitem[Torne et~al.(2024)]{rialto}
M.~Torne, A.~Simeonov, Z.~Li, et~al.
\newblock Reconciling Reality through Simulation: A Real-to-Sim-to-Real Approach for Robust Manipulation.
\newblock \emph{arXiv:2403.03949}, 2024.
\newblock \url{https://arxiv.org/abs/2403.03949}.

\bibitem[Jiang et~al.(2024)]{transic}
Y.~Jiang, C.~Wang, R.~Zhang, et~al.
\newblock TRANSIC: Sim-to-Real Policy Transfer by Learning from Online Correction.
\newblock \emph{arXiv:2405.10315}, 2024.
\newblock \url{https://arxiv.org/abs/2405.10315}.

\bibitem[Wu et~al.(2026)]{dextera}
J.~Wu, L.~Yuan, Z.~Sun, et~al.
\newblock DEXTERA: From a Single Image to Deployable Dexterous Manipulation via Real-to-Sim-to-Real.
\newblock \emph{arXiv:2609.21045}, 2026.
\newblock \url{https://arxiv.org/abs/2609.21045}.

\bibitem[Deshpande et~al.(2026)]{molmobot}
A.~Deshpande, M.~Guru, R.~Hendrix, et~al.
\newblock MolmoB0T: Large-Scale Simulation Enables Zero-Shot Manipulation.
\newblock \emph{arXiv:2603.16861}, 2026.
\newblock \url{https://arxiv.org/abs/2603.16861}.

\bibitem[Hu et~al.(2024)]{flare}
J.~Hu, R.~Hendrix, A.~Farhadi, et~al.
\newblock FLaRe: Achieving Masterful and Adaptive Robot Policies with Large-Scale Reinforcement Learning Fine-Tuning.
\newblock \emph{arXiv:2409.16578}, 2024.
\newblock \url{https://arxiv.org/abs/2409.16578}.

\bibitem[Miki et~al.(2022)]{perceptivelocomotion}
T.~Miki, J.~Lee, J.~Hwangbo, et~al.
\newblock Learning robust perceptive locomotion for quadrupedal robots in the wild.
\newblock \emph{arXiv:2201.08117}, 2022.
\newblock \url{https://arxiv.org/abs/2201.08117}.

\bibitem[Cheng et~al.(2023)]{parkour}
X.~Cheng, K.~Shi, A.~Agarwal, et~al.
\newblock Extreme Parkour with Legged Robots.
\newblock \emph{arXiv:2309.14341}, 2023.
\newblock \url{https://arxiv.org/abs/2309.14341}.

\bibitem[Ha et~al.(2024)]{umi_on_legs}
H.~Ha, Y.~Gao, Z.~Fu, et~al.
\newblock UMI on Legs: Making Manipulation Policies Mobile with Manipulation-Centric Whole-body Controllers.
\newblock \emph{arXiv:2407.10353}, 2024.
\newblock \url{https://arxiv.org/abs/2407.10353}.

\bibitem[Cheng et~al.(2024)]{navila}
A.~Cheng, Y.~Ji, Z.~Yang, et~al.
\newblock NaVILA: Legged Robot Vision-Language-Action Model for Navigation.
\newblock \emph{arXiv:2412.04453}, 2024.
\newblock \url{https://arxiv.org/abs/2412.04453}.

\bibitem[Loquercio et~al.(2021)]{flight}
A.~Loquercio, E.~Kaufmann, R.~Ranftl, et~al.
\newblock Learning High-Speed Flight in the Wild.
\newblock \emph{arXiv:2110.05113}, 2021.
\newblock \url{https://arxiv.org/abs/2110.05113}.

\bibitem[Tevet et~al.(2022)]{mdm}
G.~Tevet, S.~Raab, B.~Gordon, et~al.
\newblock Human Motion Diffusion Model.
\newblock \emph{arXiv:2209.14916}, 2022.
\newblock \url{https://arxiv.org/abs/2209.14916}.

\bibitem[Zhang et~al.(2023)]{t2mgpt}
J.~Zhang, Y.~Zhang, X.~Cun, et~al.
\newblock T2M-GPT: Generating Human Motion from Textual Descriptions with Discrete Representations.
\newblock \emph{arXiv:2301.06052}, 2023.
\newblock \url{https://arxiv.org/abs/2301.06052}.

\bibitem[Jiang et~al.(2023)]{motiongpt}
B.~Jiang, X.~Chen, W.~Liu, et~al.
\newblock MotionGPT: Human Motion as a Foreign Language.
\newblock \emph{arXiv:2306.14795}, 2023.
\newblock \url{https://arxiv.org/abs/2306.14795}.

\bibitem[Li et~al.(2023)]{omomo}
J.~Li, J.~Wu, and C.~K.~Liu.
\newblock Object Motion Guided Human Motion Synthesis.
\newblock \emph{arXiv:2309.16237}, 2023.
\newblock \url{https://arxiv.org/abs/2309.16237}.

\bibitem[Li et~al.(2023)]{chois}
J.~Li, A.~Clegg, R.~Mottaghi, et~al.
\newblock Controllable Human-Object Interaction Synthesis.
\newblock \emph{arXiv:2312.03913}, 2023.
\newblock \url{https://arxiv.org/abs/2312.03913}.

\bibitem[Araujo et~al.(2025)]{gmr}
J.~P.~Araujo, Y.~Ze, P.~Xu, et~al.
\newblock Retargeting Matters: General Motion Retargeting for Humanoid Motion Tracking.
\newblock \emph{arXiv:2510.02252}, 2025.
\newblock \url{https://arxiv.org/abs/2510.02252}.

\bibitem[Mandlekar et~al.(2023)]{mimicgen}
A.~Mandlekar, S.~Nasiriany, B.~Wen, et~al.
\newblock MimicGen: A Data Generation System for Scalable Robot Learning using Human Demonstrations.
\newblock \emph{arXiv:2310.17596}, 2023.
\newblock \url{https://arxiv.org/abs/2310.17596}.

\bibitem[Garrett et~al.(2024)]{skillgen}
C.~Garrett, A.~Mandlekar, B.~Wen, et~al.
\newblock SkillMimicGen: Automated Demonstration Generation for Efficient Skill Learning and Deployment.
\newblock \emph{arXiv:2410.18907}, 2024.
\newblock \url{https://arxiv.org/abs/2410.18907}.

\bibitem[Bharadhwaj et~al.(2024)]{gen2act}
H.~Bharadhwaj, D.~Dwibedi, A.~Gupta, et~al.
\newblock Gen2Act: Human Video Generation in Novel Scenarios enables Generalizable Robot Manipulation.
\newblock \emph{arXiv:2409.16283}, 2024.
\newblock \url{https://arxiv.org/abs/2409.16283}.

\bibitem[Jang et~al.(2025)]{dreamgen}
J.~Jang, S.~Ye, Z.~Lin, et~al.
\newblock DreamGen: Unlocking Generalization in Robot Learning through Video World Models.
\newblock \emph{arXiv:2505.12705}, 2025.
\newblock \url{https://arxiv.org/abs/2505.12705}.

\bibitem[Li et~al.(2025)]{mimicdreamer}
H.~Li, I.~Zhang, R.~Ouyang, et~al.
\newblock MimicDreamer: Aligning Human and Robot Demonstrations for Scalable VLA Training.
\newblock \emph{arXiv:2509.22199}, 2025.
\newblock \url{https://arxiv.org/abs/2509.22199}.

\bibitem[Ye et~al.(2026)]{dreamzero}
S.~Ye, Y.~Ge, K.~Zheng, et~al.
\newblock World Action Models are Zero-shot Policies.
\newblock \emph{arXiv:2602.15922}, 2026.
\newblock \url{https://arxiv.org/abs/2602.15922}.

\bibitem[Ye et~al.(2025)]{video2policy}
W.~Ye, F.~Liu, Z.~Ding, et~al.
\newblock Video2Policy: Scaling up Manipulation Tasks in Simulation through Internet Videos.
\newblock \emph{arXiv:2502.09886}, 2025.
\newblock \url{https://arxiv.org/abs/2502.09886}.

\bibitem[Kelly et~al.(2018)]{hgdagger}
M.~Kelly, C.~Sidrane, K.~Driggs-Campbell, et~al.
\newblock HG-DAgger: Interactive Imitation Learning with Human Experts.
\newblock \emph{arXiv:1810.02890}, 2018.
\newblock \url{https://arxiv.org/abs/1810.02890}.

\bibitem[Li et~al.(2026)]{rewarddagger}
R.~Li, Y.~Korkmaz, and E.~B{\i}y{\i}k.
\newblock Reward-DAgger: Robot-Gated Interactive Imitation Learning with General-Purpose Progress-Based Reward Models.
\newblock \emph{arXiv:2610.04054}, 2026.
\newblock \url{https://arxiv.org/abs/2610.04054}.

\end{thebibliography}
\end{document}